# Selective-Candidate Framework with Similarity Selection Rule for Evolutionary Optimization

Sheng Xin Zhang[a*], Wing Shing Chan[a], Zi Kang Peng[b], Shao Yong Zheng[b*], Kit Sang Tang[a]

[a]Department of Electronic Engineering, City University of Hong Kong, Kowloon, Hong Kong
[b]School of Electronics and Information Technology, Sun Yat-sen University, Guangzhou, 510006, China

**Abstract**

Achieving better exploitation and exploration capabilities (EEC) have always been an important yet challenging issue in the design of evolutionary optimization algorithm (EOA). The difficulties lie in obtaining a good balance in EEC, which is determined cooperatively by operations and parameters in an EOA. When deficiencies in exploitation or exploration are observed, most existing works consider a piecemeal approach, either by designing new operations or by altering the parameters. Unfortunately, when different situations are encountered, these proposals may fail to be the winner. To address these problems, this paper proposes an explicit EEC control method named selective-candidate framework with similarity selection rule (SCSS). M (M > 1) candidates are first generated from each current solution with independent operations and parameters to enrich the search. Then, a similarity selection rule is designed to determine the final candidate by considering the fitness ranking of the current solution and its Euclidian distance to each of these M candidates. Superior current solutions will prefer the closest candidates for efficient local exploitation while inferior ones will favor the farthest for exploration purpose. In this way, the rule could synthesize exploitation and exploration, making the evolution more effective. When applied to three classic, four state-of-the-art and four up-to-date EOAs from branches of differential evolution, evolution strategy and particle swarm optimization, significant enhancement in performance is achieved.

*Corresponding authors.
E-mail addresses: shengxinzhang@gmail.com (S. X. Zhang), zhengshaoy@mail.sysu.edu.cn (S. Y. Zheng).

## 1. Introduction

Constructed on a population basis, evolutionary optimization algorithm (EOA) explores a searching space by iteratively performing genetic operations (for evolutionary algorithms, EAs [1, 2]) or social learning processes (for swarm intelligences, SIs [3]) to generate new solutions. How these solutions are sampled, gives the feature of a particular method and determines its exploitation and exploration capabilities (EEC). For differential evolution (DE) [4-8] and evolution strategy (ES) [9], the genetic operations are mutation and crossover/recombination. While for particle swarm optimization (PSO) [10], the social learning procedures consist of the velocity and position update equations. Commonly, EEC of EOAs is indispensably controlled by the genetic operations/social learning, together with the associated parameters (e.g. mutation and crossover factors in DE, normal distribution in ES and acceleration coefficients in PSO), which cooperatively locate the sampled solutions. Since EEC is the cornerstone of evolutionary optimization [11] and has a direct impact on performance, researchers have put a lot of effort on designing appropriate exploitation and exploration schemes [12]. Existing works can be summarized under the following three categories.

(1) **EEC controlled by genetic operations/social learning.** In general, genetic operations/social learning determines the evolution direction. In this category, research works solely focus on genetic operations/social learning. Along this line, various types of operators, such as ranking-based [13], collective information-based [14] mutation, multi-objective sorting-based [15] and jumping genes-based crossover [16] were designed, favoring an exploitation or exploration trend. Fitness diversity was considered in the designs of operations [13-15]. Besides these newly designed operations, EEC were also controlled by an ensemble of multiple DE mutation strategies [17-20], a combination of different types of optimizers [21], and the memetic algorithms [22-24]. In the multialgorithm genetically adaptive method (AMALGAM) [21] and multiple offspring sampling (MOS) [23] hybrid method, the constituents compete for computational resources based on their online performance, which enhanced the exploitation capability of the unity. To preserve population diversity, [21] also introduced a diversity mechanism. In [24, 25], multiple search agents were coordinated by considering fitness distribution among individuals.

(2) **EEC controlled by parameter tuning.** Parameters control the evolution scale. In this category, researchers pursued efficient parameter tuning schemes, that included deterministic and adaptive ones. Population size is a common parameter in evolutionary optimization. In [26], population size was adaptively controlled by measuring fitness diversity. Apart from population size, extra parameters introduced into a specific algorithm may also need fine-tuning, such as the mutation and crossover factors [27] of DE. In [28, 29], fitness diversity was used in parameter controls.

(3) **EEC controlled by the combination of genetic operations/social learning and parameter tuning.**

There are also some works [30, 31] aimed at simultaneously controlling genetic operations/social learning and parameters. In [30], Mallipeddi et al. proposed to improve DE with an ensemble of parameters and mutation strategies. In [31], Wang et al. proposed to use three different mutation strategies combined with three different pairs of control parameters to generate solutions for selecting the fittest. These methods struck a balance between exploitation and exploration using two steps. The first step maintained a mutation strategy pool with diverse searching characteristics while the second step emphasized exploitation by fitness-based reward [30] or greedy pre-selection [31]. However, there were some issues that may hinder the performance. On the one hand, both methods were greedy and there was no explicit mechanism to remedy premature convergence. While on the other hand, multiple candidates were evaluated for each current solution [31], resulting in a higher total evaluation cost.

With regards to the saving of function evaluations, in the computationally expensive scenario, surrogate models, such as Kriging [32] and support vector machines (SVM) [33] were usually adopted to predict the fitness of multiple samples and select the best approximation. In [20], a cheap surrogate model was proposed to filter the densest solution from multiple samples, which were generated by multiple operators. However, due to the greedy nature, these methods may encounter difficulties when solving problems that require high population diversity.

In this paper, we propose a selective-candidate framework with similarity selection rule (SCSS), which simultaneously considers the operations (i.e. evolution direction) and parameters (i.e. evolution scale) that affect the generation of candidates while addressing the issues in category (3) and surrogate-based methods. The features, motivations and contributions of SCSS are summarized as follows.

1) SCSS first generates $M$ ($M > 1$) candidates for each current solution by $M$ independent reproduction procedures. Afterwards, one of them will become the final candidate for each current solution based on a selective rule. The big challenge here is that it should be effective and efficient. On the one hand, it is required to provide a potentially excellent candidate with balanced EEC for the next generation, while on the other hand, it should not involve objective function evaluation which requires additional cost. To resolve these issues, a similarity selection (SS) rule based on fitness ranking and Euclidian distance information is designed to strike a balanced EEC while avoiding evaluation of all the candidates.
2) SCSS also considers the fitness ranking of the population, which provides relative location information of individuals. For superior current solutions, the closest candidate measured by Euclidian distance in solution space will be selected as the final candidate for local search (exploitation) purpose. While for inferior ones, the farthest candidate is favored for basin-jumping (exploration) purpose.
3) Based on the above design, the proposed SCSS framework is expected to meet the challenge in 1) and

enhance the performance. The main contributions of this work are summarized as follows.

a) Different algorithms may be suitable for solving different optimization problems [34-36]. This study provides a generic method that is readily applied to different types of EOAs.

b) The proposed method provides an explicit EEC control paradigm based on fitness and Euclidian distance measures, which is straight-forward, simple and easy-understanding.

c) Extensive study shows that the proposed method achieves a balanced EEC and consequently demonstrates remarkable performance enhancement of several start-of-the-art and top algorithms available in the literature [37-45]. In addition, its working mechanism, benefits and real-world applications are also presented and analyzed.

The rest of this paper is organized as follows. Section 2 describes the proposed framework. Section 3 presents the experimental study and relevant discussions, while section 4 concludes this paper.

## 2 Proposed Method

### 2.1 Motivations

Generally, the procedures[1] for EAs/SIs can be summarized as **Algorithm 1.** It is common that one candidate is generated from a current solution based on the reproduction procedure. However, due to the stochastic process in operations and randomness in parameters, it cannot be guaranteed that the candidate obtained is located within promising search areas. If the reproduction procedure repeats, candidates from the same current solution are likely to be different, bringing up various building blocks, resulting in different searching performance. This is not only observed in classic EAs and SIs, but also in many of their variants. To alleviate the possible adverse effect from randomness and to improve the performance, we propose a generic selective-candidate framework with similarity selection rule (SCSS). To locate different search regions, SCSS first generates $M$ ($M > 1$) candidates with $M$ independent reproduction procedures. To save the function evaluation cost while maintaining good exploitation and exploration trade-off, a similarity selection (SS) rule based on similarity in both fitness and decision spaces is then proposed to determine one final candidate.

---

**Algorithm 1. General Procedures of EAs and SIs**

---

1: Initialize population $\boldsymbol{X} = \{\boldsymbol{x}_1, \boldsymbol{x}_2, \ldots, \boldsymbol{x}_{NP}\}$;

2: **While** the stopping criteria are not met **Do**

3: Determine the control parameters $CP$ for genetic operations/social learning;

4: Produce a new population $\boldsymbol{Y}$ via genetic operations/social learning on $\boldsymbol{X}$;

5: Evaluate the fitness of $\boldsymbol{Y}$;

[1] For brevity, a review of three typical algorithms, DE, ES and PSO is presented in the supplementary file.

6: Select solutions as new $\boldsymbol{X}$ from $\boldsymbol{X}\cup\boldsymbol{Y}$ to enter next iteration.

7: **End While**

----------------------------------------------------------------------------------------------------

### 2.2 SCSS Framework

The pseudo-code of the proposed SCSS framework is presented in **Algorithm 2**, which consists of two components, i.e. multiple candidates generation and similarity selection (SS) rule.

#### 2.2.1 Multiple Candidates Generation

As seen from Algorithm 2, the SCSS framework performs $M$ independent reproductions with $M$ sets of independent parameters (i.e. evolution scale) and operations (i.e. evolution direction) (lines 5-7). Thus, for each current solution $\boldsymbol{x}_i$, it owns a pool of candidate $\boldsymbol{y}_i^m$ $\{m = 1, 2, \ldots, M\}$. One solution $\boldsymbol{y}_i$ is then selected from the corresponding $M$ candidates for each $\boldsymbol{x}_i$ by SS rule (lines 14 and 18), as a result, the actual parameters in use are recorded (lines 15 and 19).

----------------------------------------------------------------------------------------------------

**Algorithm 2. SCSS Framework**

----------------------------------------------------------------------------------------------------

1: Initialize population $\boldsymbol{X} = \{\boldsymbol{x}_1, \boldsymbol{x}_2, \ldots, \boldsymbol{x}_{NP}\}$;

2: **While** the stopping criteria are not met **Do**

3: Determine the fitness ranking $rank(i)$ of each individual $i\{i = 1, 2, \ldots, NP\}$; // *fitness ranking for SS rule*

----------------------------- **Multiple Candidates Generation** -------------------------

4: **For** $i = 1$: $NP$

5: **For** $m = 1$: $M$

6: Determine the control parameters $CP^m=\{cp_1^m, cp_2^m, \ldots, cp_{NP}^m\}$ for genetic operations/social learning, following the original design of the baseline;

7: Produce new solution $\boldsymbol{y}_i^m$ via genetic operations/social learning on $\boldsymbol{x}_i$;

8: Calculate $dist_i^m$ = Euclidian distance $(\boldsymbol{y}_i^m, \boldsymbol{x}_i)$; // *similarity calculation for SS rule*

9: **End For**

10: **End For**

----------------------------------------------------------------------------------------------------

------------------------ **Similarity Selection Rule** -----------------------------

11: **For** $i = 1$: $NP$

12: **If** $rand_i(0,1) > rank(i)/NP$

13: $index = \arg\min_{m\in\{1,2,\ldots,M\}} (dist_i^m)$;

14: $\boldsymbol{y}_i = \boldsymbol{y}_i^{index}$;

15: $cp_i = cp_i^{index}$;

16: **Else**

17: $index = \underset{m \in \{1,2,\ldots,M\}}{\arg\max}\,(dist_i^m)$;

18: $\boldsymbol{y}_i = \boldsymbol{y}_i^{index}$;

19: $cp_i = cp_i^{index}$;

20: **End If**

21: **End For**

---------------------------------------------------------------------------------------------------

22: Evaluate the fitness of $\boldsymbol{Y}$;

23: Select solutions as new $\boldsymbol{X}$ from $\boldsymbol{X} \cup \boldsymbol{Y}$ to enter next iteration.

24: **End While**

-----------------------------------------------------------------------------------------------------------------------------------

### 2.2.2 Similarity Selection Rule

Apparently, the major challenge in the SCSS framework is how to determine the final competitor from $M$ candidates. On one hand, the selective rule should be effective to bring in performance enhancement, while on the other hand, it should be efficient to reduce the computational load.

Hence, we propose a similarity selection (SS) rule, as given in **Algorithm 2** (lines 11-21). The rule simultaneously considers the fitness ranking information $rank(i)$ of current solution $\boldsymbol{x}_i$ and its Euclidian distance $dist_i^m$ to each of the $M$ candidates $\boldsymbol{y}_i^m$, which is defined as

$$dist_i^m = \sqrt{\sum_{j=1}^{D} (y_{i,j}^m - x_{i,j})^2} \ ,$$

where $D$ is the number of decision variables.

By adjusting SS, the amount of exploitation and exploration can be directly controlled. For instance, favoring candidates closest to the current solutions are exploitative while preferring the ones farthest to the current solutions could encourage exploration.

However, it should be remarked that the appropriate choice of SS for a specific algorithm is dependent on the EEC of the given algorithm. For illustration purposes, assume that the EEC is represented by a searching radius (SRAD). A larger SRAD implies a more explorative characteristic, and vice versa.

For a very explorative optimizer (named Optimizer 1), the large SRAD facilitates a more random-like search with little risk suffered from local optima. However, this large SRAD would make the current individuals hard to refine. The reason is that the size of search space increases exponentially with the number of dimensions, thus, the current individuals, as preserved by evolutionary algorithms, are relatively scarce. Over-large SRAD is likely to visit huge useless areas and break up useful genetic materials.

For a very exploitative optimizer (named Optimizer 2), the small SRAD would make the solutions focus more on local searches and are unlikely to explore new materials. It is difficult for them to move from one basin to another, which is important when addressing complicated multi-modal problems.

For a balanced optimizer (named Optimizer 3) with an appropriate SRAD, the SRAD could vary among individuals for a better design. For example, when considering the fitness, with superior solutions in promising areas, a smaller and appropriate SRAD may lead to better exploitation. While inferior solutions are likely to be located near the peaks. A larger and appropriate SRAD could force them to explore nearby basins.

The above explanation is based on the premise that a smaller SRAD has more chances of providing better solutions while a larger SRAD has relatively fewer chances but higher population diversity, which is confirmed by our experiments (see Section 3.3.3 and Fig.5).

Regarding different cases: 1) for Optimizer 1, the SRAD should be reduced to concentrate the search; 2) for Optimizer 2, the SRAD should be enlarged to encourage exploration to new searching areas; and 3) for Optimizer 3, different searching tasks should be assigned to solutions with different potentials. On the one hand, new best solutions are likely to be located in the area near the top-ranked solutions in the context of a continuous landscape. To achieve a better efficiency in exploitation, closest candidates of superior solutions are considered, targeting steady improvements for promising areas. On the other hand, farthest candidates of inferior solutions are preferred, aiming for better exploration.

In view of the above, two SS schemes are proposed as follows:

**Scheme 1**: **If** $rank(i) \leq \text{ceil}(NP \times GD)$

Select the closest candidate from $\boldsymbol{y}_i^m$ $\{m = 1, 2, \ldots, M\}$ for individual $\boldsymbol{x}_i$;

**Else**

Select the farthest candidate from $\boldsymbol{y}_i^m$ $\{m = 1, 2, \ldots, M\}$ for individual $\boldsymbol{x}_i$;

**End If**

**Scheme 2**: **If** $rand_i(0,1) > rank(i)/NP$

Select the closest candidate from $\boldsymbol{y}_i^m$ $\{m = 1, 2, \ldots, M\}$ for individual $\boldsymbol{x}_i$;

**Else**

Select the farthest candidate from $\boldsymbol{y}_i^m$ $\{m = 1, 2, \ldots, M\}$ for individual $\boldsymbol{x}_i$;

**End If**

where $rank(i) \in \{1, 2, \ldots, NP\}$ is the fitness ranking of individual $\boldsymbol{x}_i$ and $rank(i)=1$ is the best. *ceil (.)* is a ceiling function. $rand_i(0,1)$ is a uniformly distributed random number within (0,1) for individual $\boldsymbol{x}_i$ $\{i = 1, 2, \ldots, NP\}$.

In Scheme 1, the proportion of top individuals preferring the closest candidates is controlled by a greedy degree parameter $GD$ in the range [0,1]. Specifically, the superior $GD \times 100\%$ selects the nearest candidates while the inferior $(1 - GD) \times 100\%$ portion selects the farthest candidates. The larger the $GD$ value is, the more exploitative Scheme 1 becomes.

In Scheme 2, higher ranked individuals are associated with higher probabilities in using the closest candidates, while lower ranked ones are likely to utilize the farthest candidates. One of the advantages is that Scheme 2 is parameterless. As shown later in Section 3, Scheme 2 works well for most of the advanced EA and SI variants.

Based on Algorithm 2, the SCSS variants for existing EAs and SIs can be easily implemented. Examples on the work flow of three SCSS variants, namely SCSS-DE, SCSS-ES and SCSS-PSO for the classic DE, ES, and PSO are given in Algorithms S1, S2 and S3 in the supplementary file, respectively.

### 2.2.3 Time Complexity

This subsection discusses the time complexity of the proposed method. Consider DE as an example, its time complexity is $O(NP \cdot D \cdot Gen_{\max})$, where $NP$ is population size, $D$ is the number of decision variables of the problem and $Gen_{\max}$ is the maximum number of generations. In SCSS-DE, the complexity of fitness ranking and Euclidian distance calculation for each generation are $O(NP \cdot \log_2 NP)$ and $O(M \cdot NP \cdot D)$, respectively. Besides, the complexity of $M$ reproductions is $O(M \cdot NP \cdot D)$. Since $\log_2 NP << D$, the overall complexity is $O(M \cdot NP \cdot D \cdot Gen_{\max})$. As investigated in Section 3, $M = 2 << NP$ is sufficient for advanced DEs, such as the JADE [37] and L-SHADE [41] algorithms. Thus, the complexity of advanced SCSS-DEs remains as $O(NP \cdot D \cdot Gen_{\max})$.

## 3 Simulation

In this section, the effectiveness of the proposed SCSS framework and its working mechanism are investigated through comprehensive experiments conducted using the CEC2014 [47] and CEC2017 [48] benchmark function sets. Each function set consists of 30 functions with diverse mathematic characteristics, such as unimodal, multimodal, hybrid and composition. Since the CEC function suits have bounded constraints, to make the comparison fair, the constraint handling technique adopted in the SCSS variants is kept the same as the corresponding baselines. The solution error value, defined as $f(x) - f(x^*)$, is used to measure the performance of the compared algorithms, where $f(x)$ is the smallest fitness obtained after $10^4 \times D$ function evaluations and $f(x^*)$ is the fitness of the global optimal $x^*$. In line with [47, 48], solution error values smaller than $10^{-8}$ are considered as zero. For each test function, 51 independent runs are performed, while the mean and standard deviations of the solution error values are reported. In order to draw statistically sounded conclusions, Wilcoxon signed-rank test [49] with 5% significance level is applied to compare the performance. The symbols "－", "＝" and "+" represent that the baseline algorithms perform significantly worse than, similar to or better than the corresponding SCSS variants, respectively. The significant ones are marked in **bold**.

### 3.1 Performance Enhancement of Classic EAs and SIs

The proposed SCSS framework is first integrated with three classic EAs and SIs, i.e. DE and ES from EA family and PSO from SIs. Performance of the resulting variants, SCSS-DE, SCSS-ES and SCSS-PSO are compared with the baseline algorithms, respectively.

The parameter settings for the compared algorithms are summarized as follows:

DE and SCSS-DE: $NP = 100$, $F = 0.7$, $CR = 0.5$;

ES and SCSS-ES: $\mu = 25$, $\lambda = 100$, intermediate recombination is used;

PSO and SCSS-PSO: $NP = 20$, $w = 0.9$, $c_1 = 2.0$, and $c_2 = 2.0$;

In addition, regarding the SS rule, Scheme 1 with $GD = 1$ and $M = 2$ is adopted in the three SCSS variants. These settings are based on the experimental findings given later in Section. 3.3. Comparison results on 30-$D$ and 50-$D$ CEC2014 functions are summarized in Fig.1.

As observed in Fig. 1, the effectiveness of the proposed SCSS framework on all the considered algorithms is verified. Of the 180 cases in total, SCSS variants win in 125 (=21+26+15+22+27+14) and only lose in one. Specifically, in the 30-$D$ cases, SCSS-DE and SCSS-ES perform significantly better than their corresponding baselines on 21 and 26 functions and lose on one and no function, respectively. SCSS-PSO wins PSO on 15 functions and ties on 15 functions. In the 50-$D$ case, SCSS-DE, SCSS-ES, and SCSS-PSO win the baselines on 22, 27 and 14 functions, respectively, while the rest are tie. It is noted that, since the classic algorithms use fixed parameter settings, these performance improvements are attributed to the control of the randomness of the reproduction operations by SCSS, such as the random selection of parents for mutation and dimension-wise crossover in DE. In summary, the proposed SCSS framework significantly enhances the performance of these basic algorithms.

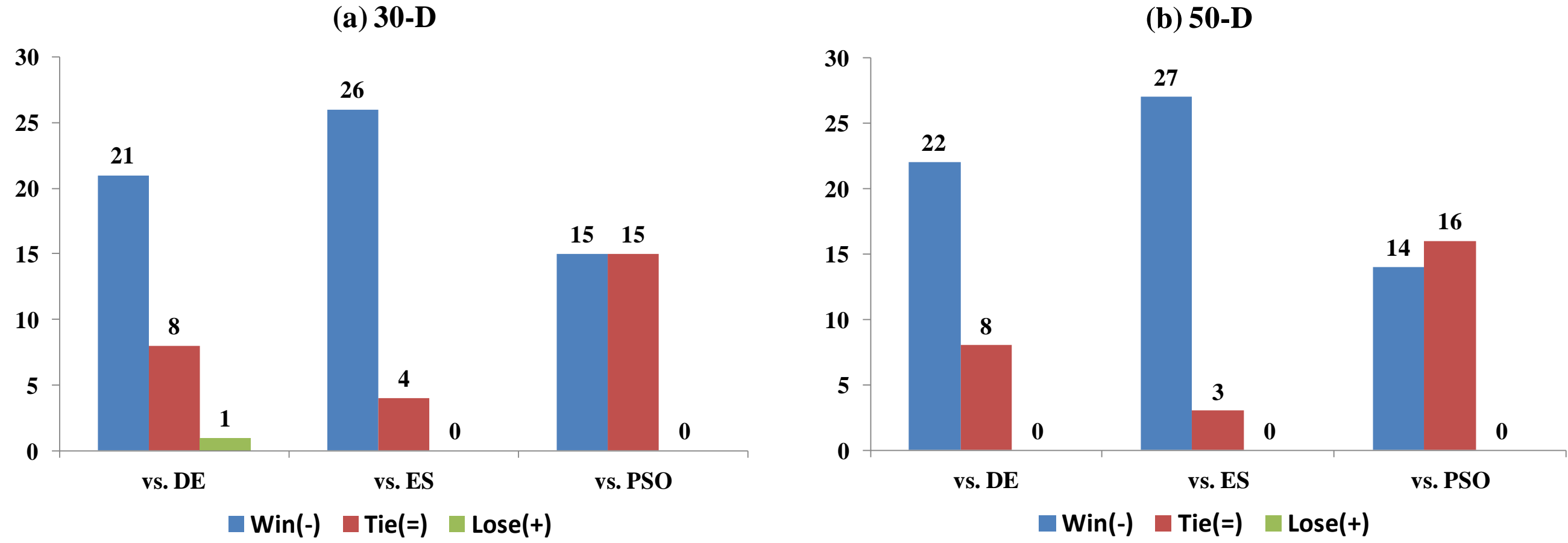

Fig.1 Comparison results of three SCSS-based classic algorithms with the baselines on CEC2014 test functions: (a) 30-$D$, (b) 50-$D$. Scheme 1 with $GD = 1$ and $M = 2$ for all the three SCSS variants.

### 3.2 Performance Enhancement of Advanced EAs and SIs

Due to the efforts by EA and SI researchers, performance of the classic algorithms has been greatly improved by many advanced variants. Thus, it is essential to investigate whether our proposed method could also further improve these algorithms. For demonstration, SCSS is incorporated into four advanced baselines, namely JADE [37], SHADE [38], CMA-ES [39] and LIPS [40]. Parameter settings for the compared algorithms are set to be the same as those recommended in their original literature. Additionally, for the SCSSs, Scheme 2 is utilized as the SS rule in SCSS-JADE, SCSS-SHADE and SCSS-LIPS, while Scheme 1 with $GD$ = 0 is applied for SCSS-CMA-ES. The reproduction times $M$ is set to 2 for SCSS-JADE and SCSS-SHADE, 4 for SCSS-LIPS and 5 for SCSS-CMA-ES. These settings are the best, as indicated later by the parameter sensitivity analyses in Section. 3.3.

The experimental results on 30-*D* and 50-*D* CEC2014 functions are shown in Table S1 and Table S2, respectively, in the supplementary file and further summarized in Fig. 2.

As observed from Fig. 2, SCSS also exhibits remarkable improvements on the advanced algorithms. Out of the 240 cases in total, SCSS wins in 134 (=14+14+17+23+16+11+13+26) and loses in just 17 (=1+0+5+2+1+0+5+3). More specifically, for the advanced DEs, i.e. JADE and SHADE, SCSS improves their performance on 55 functions and is inferior on only 2 functions. For CMA-ES, SCSS wins in 17 and 13 cases and loses in 5 cases on the 30-*D* and 50-*D* functions, respectively. For the advanced PSO algorithm, i.e. LIPS, SCSS-LIPS is superior on more than 20 functions and inferior on far fewer functions in both 30-*D* and 50-*D* cases.

Considering the diverse mathematical properties of the test functions, it can be concluded that SCSS consistently works well on various types of functions, including unimodal, multimodal, hybrid and composition.

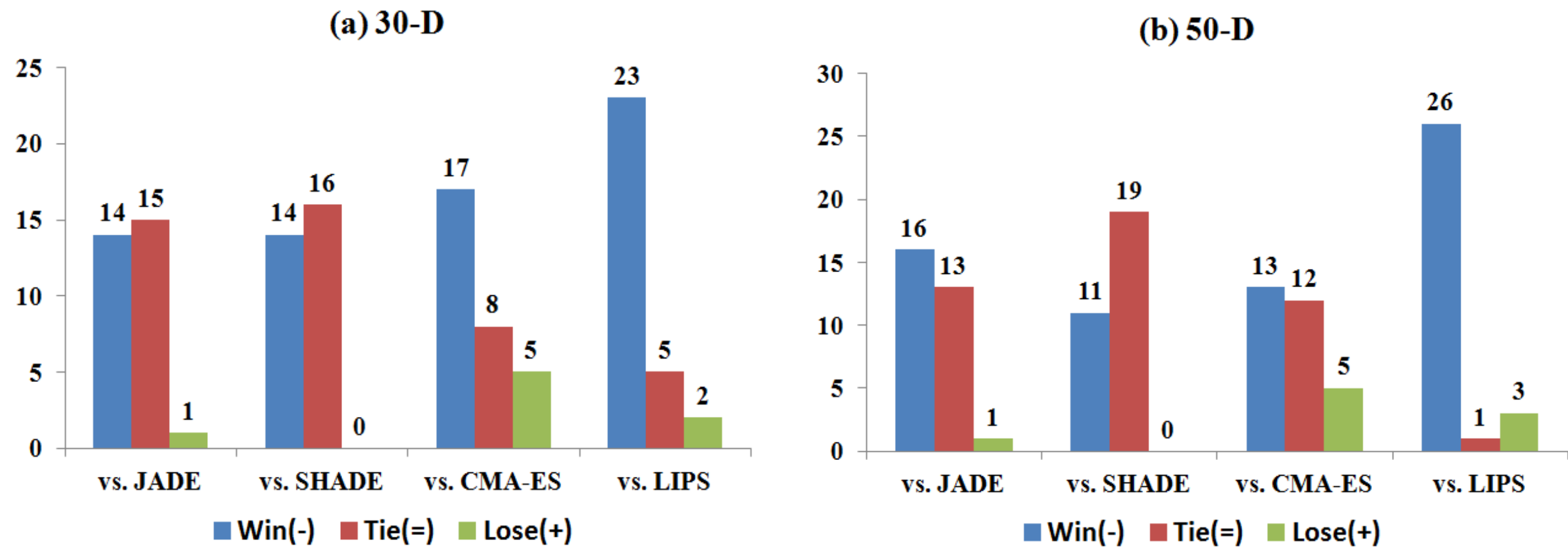

Fig.2 Comparison results of four SCSS-based advanced algorithms with the baselines on CEC2014 test functions: (a) 30-*D*, (b) 50-*D*. Scheme 2 is utilized in SCSS-JADE, SCSS-SHADE and SCSS-LIPS, while Scheme 1 with $GD$ = 0 is applied for SCSS-CMA-ES. The reproduction times $M$ is set to 2 for SCSS-JADE and SCSS-SHADE, 4 for SCSS-LIPS and 5 for SCSS-CMA-ES.

### 3.3 Working Mechanism of SS Rule

#### 3.3.1 Influence of SS rule on the performance of SCSS

The performance sensitivity of SCSS to the SS rule is first investigated. Performance of seven SCSSs, i.e. SCSS-DE, SCSS-ES, SCSS-PSO, SCSS-JADE, SCSS-SHADE, SCSS-CMA-ES and SCSS-LIPS with different SS rules (i.e. Scheme 1 with six *GD* values, i.e. 0, 0.2, 0.4, 0.6, 0.8, 1 and Scheme 2) are compared with those of the baseline algorithms, respectively. The *M* value for all the SCSS variants in this experiment is set as 2. The completed comparison results "-/=/+" are given in Table S3 in the supplementary file, while Fig. 3 presents the P-N values (defined as the number of "－" minus the number of "+") as a summary.

From Fig. 3, the followings can be observed:

(1) For the classic algorithms, including DE, ES, and PSO, SCSS variants adopting larger *GD* values perform better than those with smaller ones. The reason is that classic algorithms are usually explorative and deficit in exploitation (the case of Optimizer 1 in Section 2.2.2). Large *GD* values could encourage exploitation to remedy the blindness of the search. Small *GD* values, such as *GD*=0, make the algorithms even more explorative and deteriorate the performance, as can be observed from Fig. 3.

(2) For the advanced algorithms, Scheme 2 is the best choice for SCSS-SHADE and SCSS-LIPS and the third best choice for SCSS-JADE. Also, for SCSS-JADE and SCSS-SHADE, the performance of SCSSs with Scheme 1 significantly degenerates when *GD* is too large (*GD* = 1) or too small (*GD* = 0). This is because JADE and SHADE themselves maintain relatively balanced EEC (the case of Optimizer 3 in Section 2.2.2). *GD*=1 would over-emphasize exploitation and make the algorithms too greedy while an over-explorative setting *GD* = 0 may deteriorate the performance on test functions which need more exploitation.

(3) For SCSS-CMA-ES, Scheme 1 with *GD* = 0 achieves the best performance, indicating that the original CMA-ES (the case of Optimizer 2 in Section 2.2.2) needs more exploration for performance enhancement. This observation is consistent with the statements in some CMA-ES literature, such as IPOP-CMA-ES [46] that CMA-ES could benefit from enhanced exploration capability when solving difficult CEC benchmarks. The restart mechanism proposed in IPOP-CMA-ES enhances the population diversity. To investigate the effectiveness of SCSS on IPOP-CMA-ES, we also implemented the SCSS-IPOP-CMA-ES algorithm. According to our experiment, the optimal SS rule for SCSS-IPOP-CMA-ES goes to Scheme 2. Comparative results of SCSS-IPOP-CMA-ES with IPOP-CMA-ES are presented in Table S4 in the supplementary file. It can be seen that SCSS-IPOP-CMA-ES performs slightly better than IPOP-CMA-ES in 30-D but worse in 50-D with the “win/tie/lose” counts of “5/23/2” and “4/20/6” respectively. For most of the functions, the two algorithms perform similarly with the superiority of SCSS being less obvious. There may be two reasons: 1) The restart mechanism successfully maintains population diversity and becomes the dominate component; 2) the restart mechanism may be inappropriate for the SCSS method since it was specifically designed for IPOP-CMA-ES. Nevertheless, designing new restart conditions is a topic that deserves future investigations.

In conclusion, the choice of the best SS rule depends on the EEC of the baselines while Scheme 2 consistently performs significantly better than or similar to the baselines. As a design rule of thumb, for an optimizer with relatively balanced EEC, Scheme 2 is recommended.

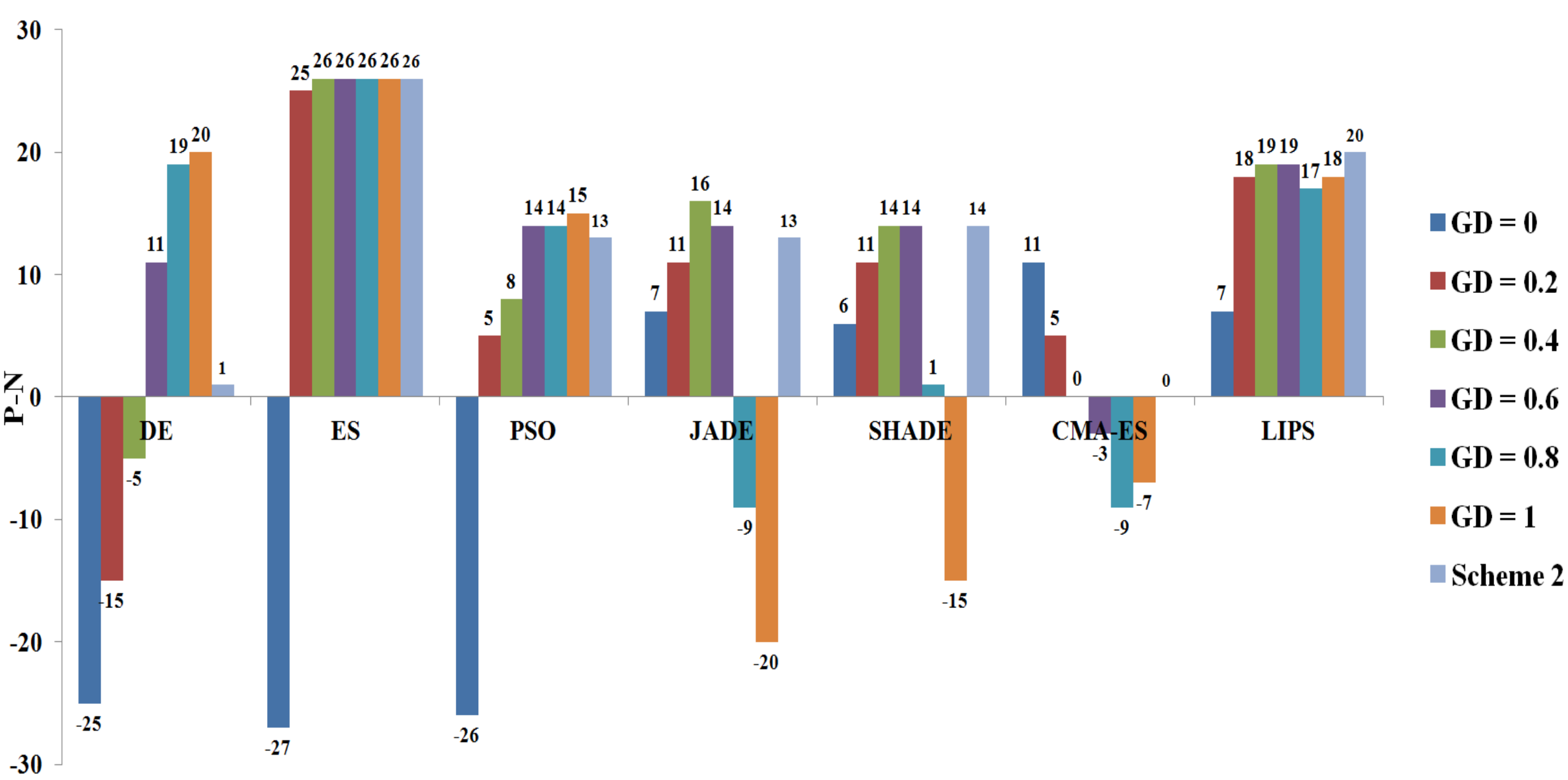

Fig.3 P-N values of SCSS variants with different SS rules against the baselines on 30-D CEC2014 test functions. (P-N value = the number of functions that SCSS variant outperforms the baseline − the number of functions that SCSS variant loses to the baseline).

### 3.3.2 Behavior of SS rule

In the proposed SCSS framework, the selection of the closest or farthest candidates is conducted based on the fitness ranking of the current solutions. In this way, SCSS adjusts the level of exploration/exploitation according to their potential. In the experiment conducted on SCSS-DE ($GD = 1$) and SCSS-SHADE (Scheme 2), SS rule is compared with a randomly selecting (RS) manner (i.e. selecting manner in the baseline algorithm). The total distance *TD* between the selected candidates and the current solutions against the rank on 30-*D* functions F5 and F13 in the median run is shown in Fig. 4.

From Fig. 4, we have the following observations: 1) on the explorative DE, SS enhances the exploitation on all ranks, resulting in smaller *TD* values than that of RS; 2) on SHADE with relatively balanced EEC, for ranks smaller than $NP/2 = 50$, SS yields smaller *TD* compared to RS, resulting in more exploitation. While for ranks larger than 50, it is the opposite case; 3) on SHADE, for RS, *TD* varies little with the rank but *TD* significantly increases with the rank for SS. Since the searching radius SRAD can be roughly calculated as $TD/Gen_{\max}$, where $Gen_{\max}$ is the maximum number of generations and is the same for both SHADE and SCSS-SHADE, $SRAD \propto TD$ . This means that SRAD increases with the rank in SS while it is the same in RS. In other words, SS is a finer strategy; 4) the smaller *TD* values of SHADE compared to that of DE reveal that SHADE is more exploitative than DE. Therefore, unlike the case in SCSS-DE, enlarging *GD* in SCSS-SHADE may make the algorithm over-exploitative and deteriorate the performance, which is also observed from Fig. 3.

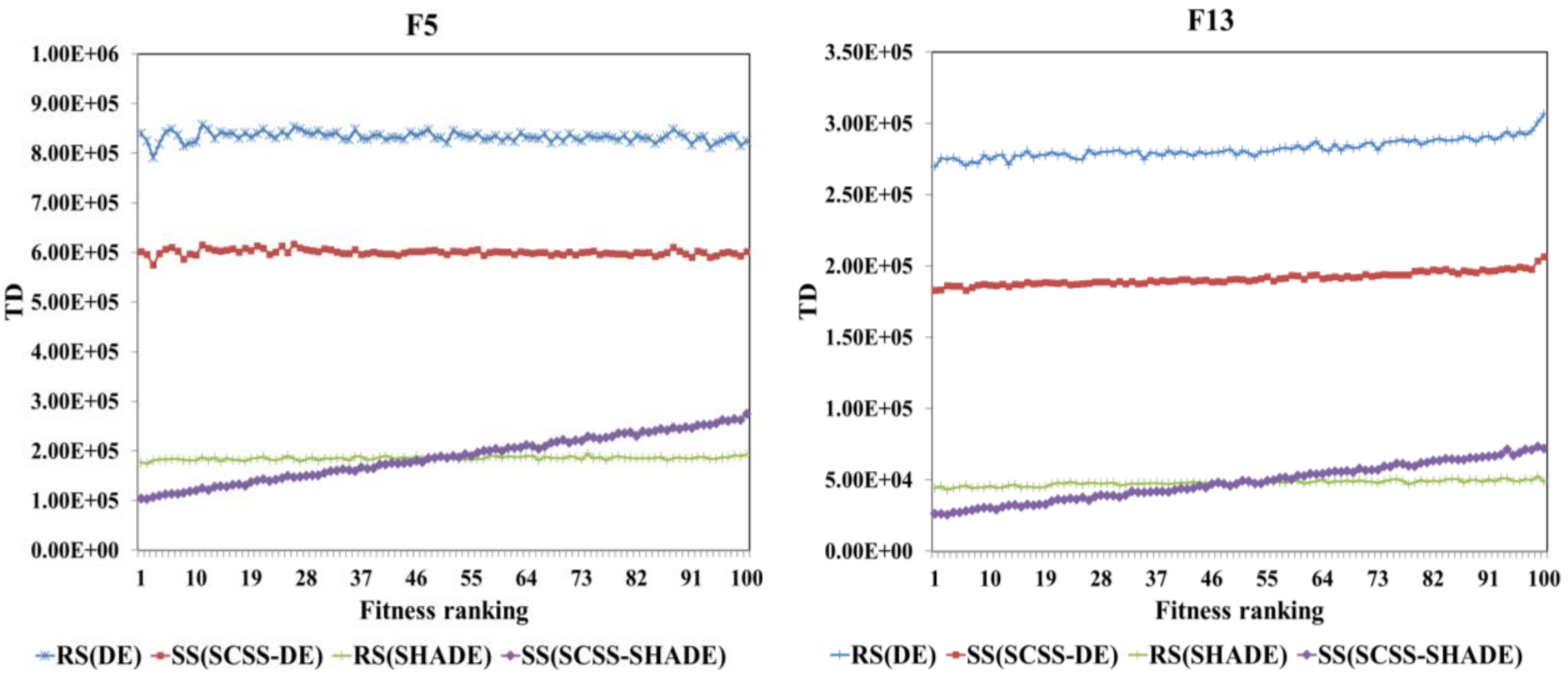

Fig.4 TD against the rank on 30-D CEC2014 functions F5 and F13. (The similar phenomena can be observed on all the CEC functions)

### 3.3.3 Benefit of SS rule

To further demonstrate the benefit of SS (Scheme 2), it is compared with the following three variants:

*Variant-oppo*: An opposite version of Scheme 2 is defined as follows:

**If** $rand_i(0,1) > rank(i)/NP$

Select the farthest candidate from $\boldsymbol{y}_i^m$ $\{m = 1, 2, \ldots, M\}$ for individual $\boldsymbol{x}_i$;

**Else**

Select the closest candidate from $\boldsymbol{y}_i^m$ $\{m = 1, 2, \ldots, M\}$ for individual $\boldsymbol{x}_i$;

**End If**

*Variant-Meval*: Scheme 2 is replaced with true function evaluations. Specifically, for each current solution, $M$ candidates are evaluated and the fittest one is selected as offspring, like CoDE [31].

*Variant-CSM*: Instead of using Scheme 2, the cheap surrogate model (CSM) proposed in [20] is used to determine the offspring from $M$ candidates.

For a direct comparison, other settings are unaltered for experiments conducted with JADE [37]. From Table 1 and Table S5, the results are summarized as follows.

(1) SCSS-JADE exhibits better performance than Variant-oppo. In addition, comparing Table S1 with Table S5, it is also observed that Variant-oppo performs significantly worse than the baseline, concluding that the opposite version is an inappropriate selective rule. This confirms the illustrations given in Section 2.2.2.

(2) SCSS-JADE performs better than Variant-Meval. This can be explained by the fact that, in Variant-Meval, $M$ ($M$ = 2) function evaluations are consumed to determine each offspring per geneation and, as a result, the maximum number of iterations is reduced. (Note: The total number of evaluations are fixed.)

(3) SCSS-JADE also outperforms Variant-CSM. To have an in-depth insight into the working processes of SS and CSM, Fig. 5 plots their average prediction accuracy (PA) on thirty 30-D CEC2014 functions. The PA is calculated as the number of trials that correctly selects the fittest candidate divided by the number of total trials. From Fig. 5, we have the following observations and discussions.

1) Overall, PA varies with problems that pose different degree of difficulties.

2) For SS, exploitation part (EiP) has higher PA than the exploration part (ErP) on all the functions. This is understandable as ErP is responsible for broadening the search region.

3) Comparing EiP with CSM, it is seen that EiP has a higher PA on 24 out of the total 30 functions. As pointed out in [20], since CSM is a cheap model, it may not estimate the density exactly, especially for the highly-rotated CEC test functions.

4) Although a high PA is generally more desirable, it does not necessarily contribute to better performance on some functions. This can be confirmed by the observations on F17, F18 and F24. On these three functions, although CSM has higher PA than EiP, its performance is significantly inferior to SS (see Table S5). This is because CSM includes no mechanism for exploration while SS maintains two strategies simultaneously (i.e. superior/inferior solutions select the closest/farthest candidates) for synthesizing exploitaton and exploration purposes, respectively. The latter strategy always attemps to explore far-away areas, where new exploitation may then emerge once the offspring of inferior solutions becomes elites. For this reason, it is expected that exploration could also benefit exploitation and should work cooperatively. In fact, this has been verified by the overwhelmingly better performance of SCSS-JADE with Scheme 2 over $GD = 1$ (see Fig. 3).

5) Besides accuracy, it is noted that the SS rule has lower complexity ($O(M \cdot NP \cdot D)$) than CSM ($O(M \cdot NP^2 \cdot D)$ [20]), which is more significant with larger $NP$ value.

Table 1 Comparison results of SCSS-JADE with three variants on 30-D CEC2014 test functions

| -/=/+ | |
|---|---|
| Variant-oppo vs. SCSS-JADE | 24/5/1 |
| Variant-Meval vs. SCSS-JADE | 16/14/0 |
| Variant-CSM vs. SCSS-JADE | 18/11/1 |

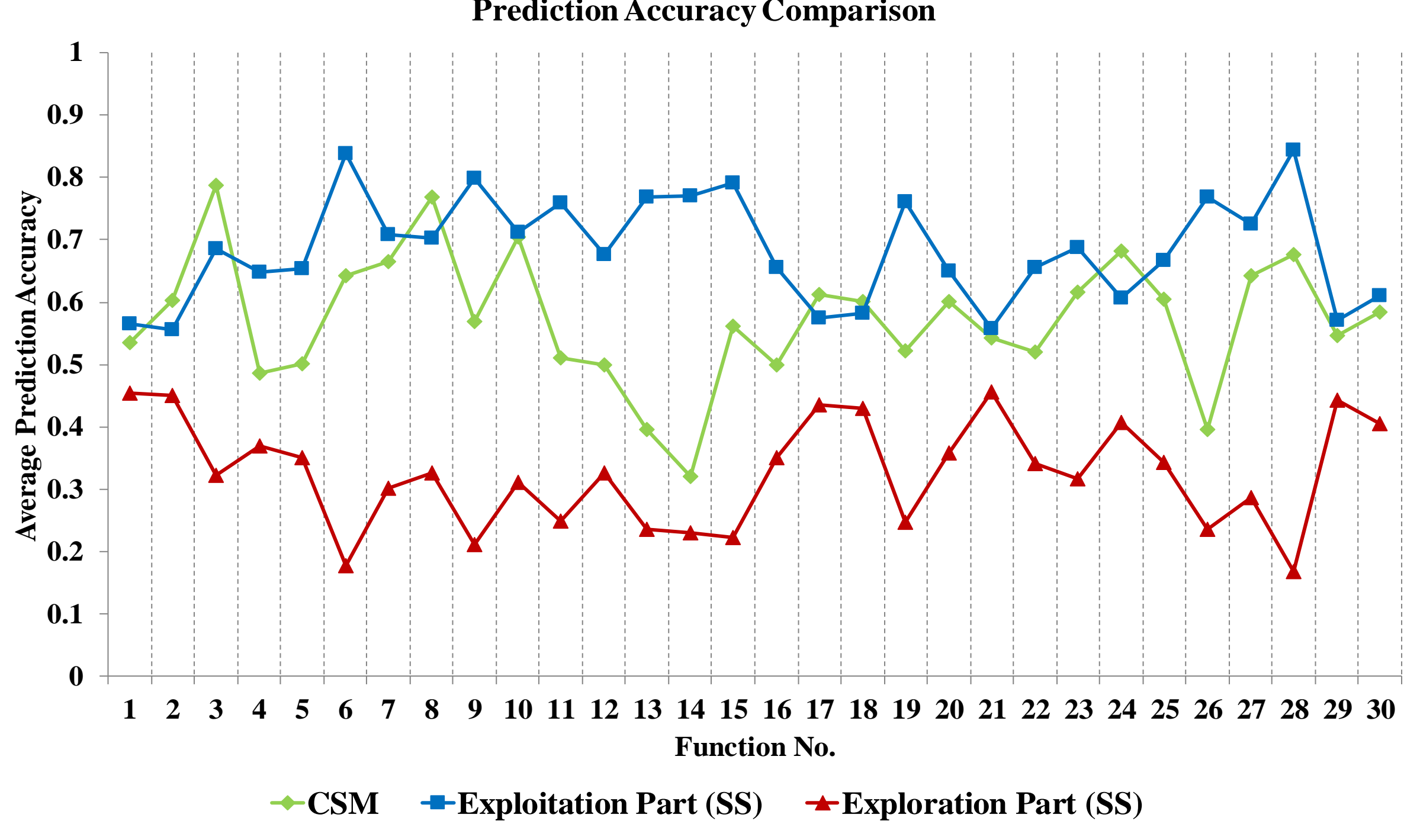

Fig. 5 Comparison of average prediction accuracy between SS and CSM on thirty 30-D CEC2014 functions.

### 3.3.4 Combined effects of operations and parameters by SS rule

The SS rule considers candidates that reveal the combined effects of operations and parameters, which makes SCSS a general framework that can be easily applied to various types of EAs and SIs. The effects of SCSS on the randomness of operations and parameters of the previously considered algorithms are summarized as follows.

(1) For the three classic algorithms DE, ES and PSO, since the parameters are fixed during the entire evolution process, SCSS reveals the effect of operations;

(2) For the advanced DEs, i.e. JADE and SHADE, except the operations, since different reproduction procedure *m* may use different *F* and *CR*, SCSS reveals their combined effects;

(3) In the advanced ES, i.e. CMA-ES, new individuals are generated from the center of best solutions by following a normal distribution. Thus, in SCSS-CMA-ES, different normal distributions are sampled in different reproduction procedures;

(4) In the advanced PSO, i.e. LIPS, SCSS uses different independently generated $\varphi_j$ in the position update equation, which is a uniformly distributed random number ranged in [0, 4.1/ *neighborhood size*] for each dimension *j* [40].

### 3.4 Performance Sensitivity to M

In SCSS, $M$ ($M > 1$) reproduction procedures should be performed. Indeed, if $M$ is set to 1, SCSS variants degenerate to baselines. The performance of SCSS is influenced by $M$. Therefore, in this subsection, SCSS variants with five different $M$ values, i.e. $M = 2$, 3, 4, 5 and 10 are compared. Except for $M$, other parameter settings are set the same as those used previously in Sections 3.1 and 3.2. Performance comparisons of the SCSS variants with the baselines on 30-*D* CEC2014 functions are summarized in Table S6 and Fig. 6. In addition, to show the dynamic performance variation with increasing $M$, the performance of the SCSS variants using adjacent $M$ settings are also compared with each other, as shown in Table S7 and Fig. 7.

It can be observed from Fig. 6 that all $M$ settings significantly improve the performance of the baselines except SCSS-JADE and SCSS-SHADE with $M = 10$.

In Fig.7, for clarity, the algorithms are divided into two categories. Category 1 includes the SCSS variants which may perform significantly better with $M > 2$ than with $M = 2$, while Category 2 lists the SCSS variants which perform similarly or even worse with increased $M$ values.

In Category 1, it is observed that the performance of DE and ES consistently improves when $M$ increases. In this paper, we only investigate cases up to $M$=10 since these classic algorithms are significantly inferior to the advanced algorithms. Moreover, increasing $M$ will increase the computational complexity of the algorithm. For CMA-ES and LIPS, SCSS variants with $M = 5$ and $M = 4$ show the best performance, respectively. It is noticed that in SCSS-CMA-ES, *GD* is set to 0, thus, larger $M$ values would make the algorithm more explorative.

In Category 2, enlarging $M$ does not bring significant performance improvements. On the contrary, it may even significantly degrade the performance, eg. $M > 4$ for JADE and SHADE, or $M > 2$ for PSO. The reason is that different from those in Category 1 (eg. DE, ES and LIPS), JADE, SHADE and PSO include elite individuals in their reproduction processes. Specifically, the top-ranked individuals used in the "current-to-*pbest*/1" mutation strategy of JADE and SHADE and the global best *gbest* used in the velocity update equation of PSO. Therefore, algorithms with too large $M$ value is potentially too greedy, making the algorithms stuck in a local optima.

Overall, it can be concluded that the appropriate $M$ value is relatively small for advanced variants.

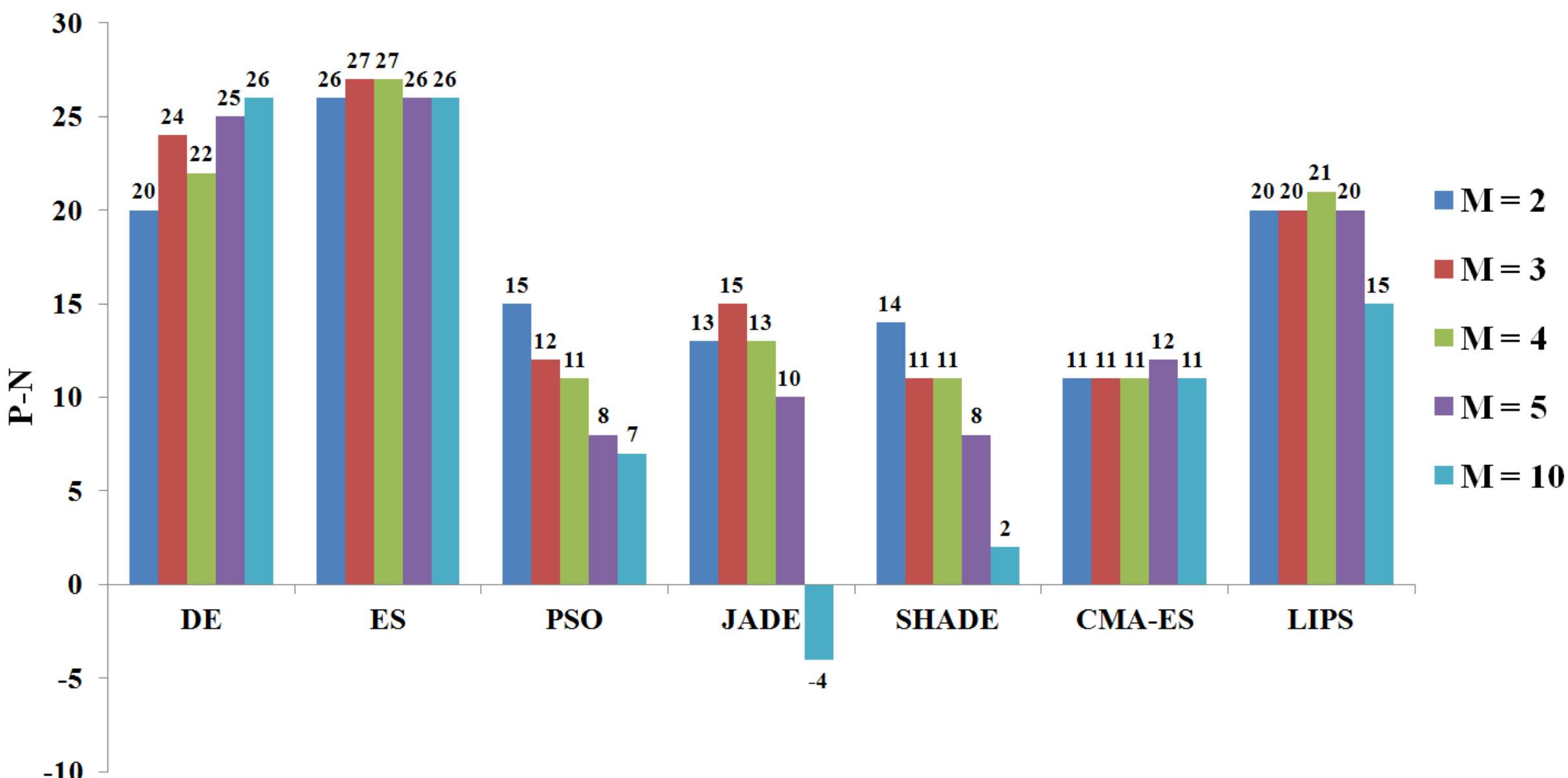

Fig.6 P-N values of SCSS variants with different M settings against the baselines. (P-N value = the number of functions that SCSS variant outperforms the baseline − the number of functions that SCSS variant loses to the baseline).

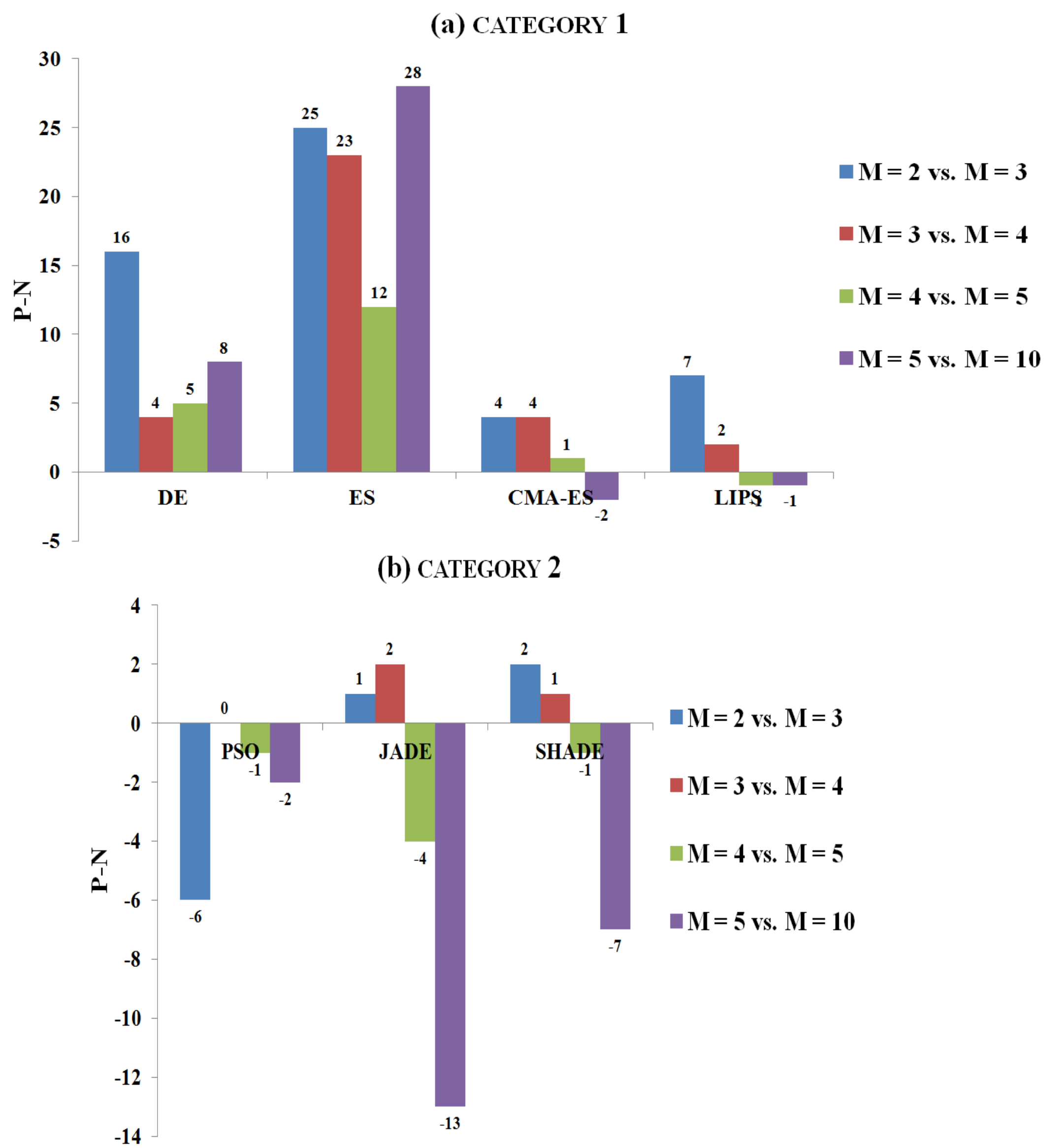

Fig.7 P-N values between SCSS variants (*A* vs. *B*) with adjacent M settings (P-N value = the number of functions that *B* outperforms *A* − the number of functions that *B* loses to *A*).

### 3.5 Application in Top Methods from CEC Competitions

From Sections 3.3 and 3.4, it can be concluded that advanced SCSS-DEs with Scheme 2 and $M$ = 2, SCSS-CMA-ES with Scheme 1($GD$ = 0) and $M$ = 5 exhibit promising performance. In this subsection, to demonstrate the flexibility, SCSS is further applied with these settings to four highly competitive algorithms from the CEC competitions. Among them, L-SHADE [41] is the winner of the CEC2014 competition, UMOEA-II [42] and L-SHADE_EpSin [43] are the joint-winner of the CEC2016 competition and jSO [44] is one of the best-performing algorithms in the CEC2017 competition. Their source codes are available at http://www.ntu.edu.sg/home/epnsugan/. Parameter settings for these top algorithms are set the same as the original literature.

As shown in Table S8, Table S9 and Fig. 8, SCSS also enhances the performance of these top methods. Out of the total 240 cases, SCSSs win in 88 (=10+9+8+7+18+10+13+13) cases and lose in 12 (=2+1+0+2+2+3+0+2) cases. Specifically, in the 30-*D* case, SCSS-L-SHADE, SCSS-UMOEA-II, SCSS-L-SHADE_EpSin, and SCSS-jSO perform significantly better than the corresponding baselines in 10, 9, 8 and 7 cases and underperform in 2, 1, 0 and 2 cases, respectively. In the 50-*D* case, the performance improvements are more significant. SCSS-L-SHADE, SCSS-UMOEA-II, SCSS-L-SHADE_EpSin and SCSS-jSO exhibit superior performance on 18, 10, 13 and 13 functions respectively and are inferior on far fewer functions.

Fig. 9 shows the convergence plot of SCSS-L-SHADE versus L-SHADE on six selected 50-*D* CEC2014 functions. As observed, SCSS-L-SHADE exhibits better convergence than L-SHADE. In conclusion, these performance enhancements indicate that the proposed SCSS framework is a better alternative for these top algorithms.

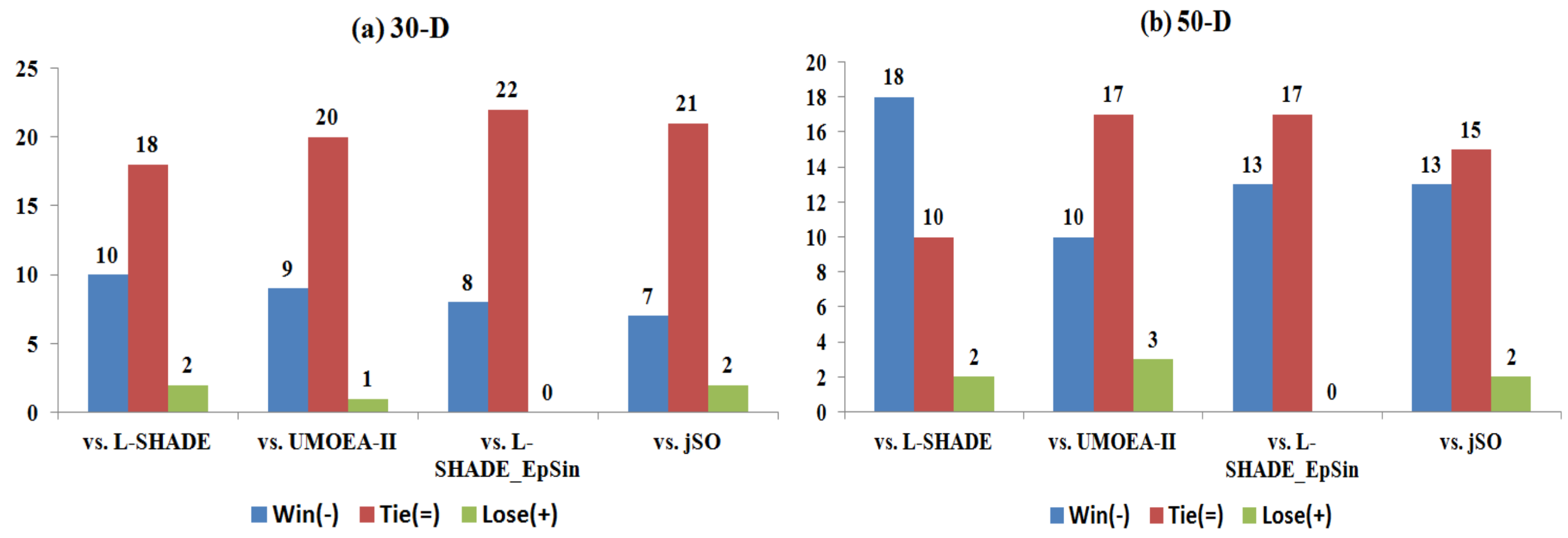

Fig.8 Comparison results of four SCSS-based top algorithms with the baselines on CEC2014 test functions: (a) 30-D, (b) 50-D.

### 3.6 Performance on CEC2017 Test Suit and Scalability Study

To assess the performance of SCSS on a wider variety of functions, in this subsection, we further test the advanced SCSS variants on the recently developed CEC2017 test suite [48]. This test suite also has 30 functions, but with several new features, such as new basic functions, graded level of linkages and rotated trap functions [48].

Parameter settings for the algorithms are the same as those used in Sections 3.2 and 3.5. Tables S10-S13 present the experimental results on 30-*D* and 50-*D* functions and Table 2 summarizes the comparison results. From Table 2, it is clear that SCSS also significantly improves the performance of the baselines on the CEC2017 functions. In the total of 480 cases, SCSS wins in 225, ties in 240 and loses in 15.

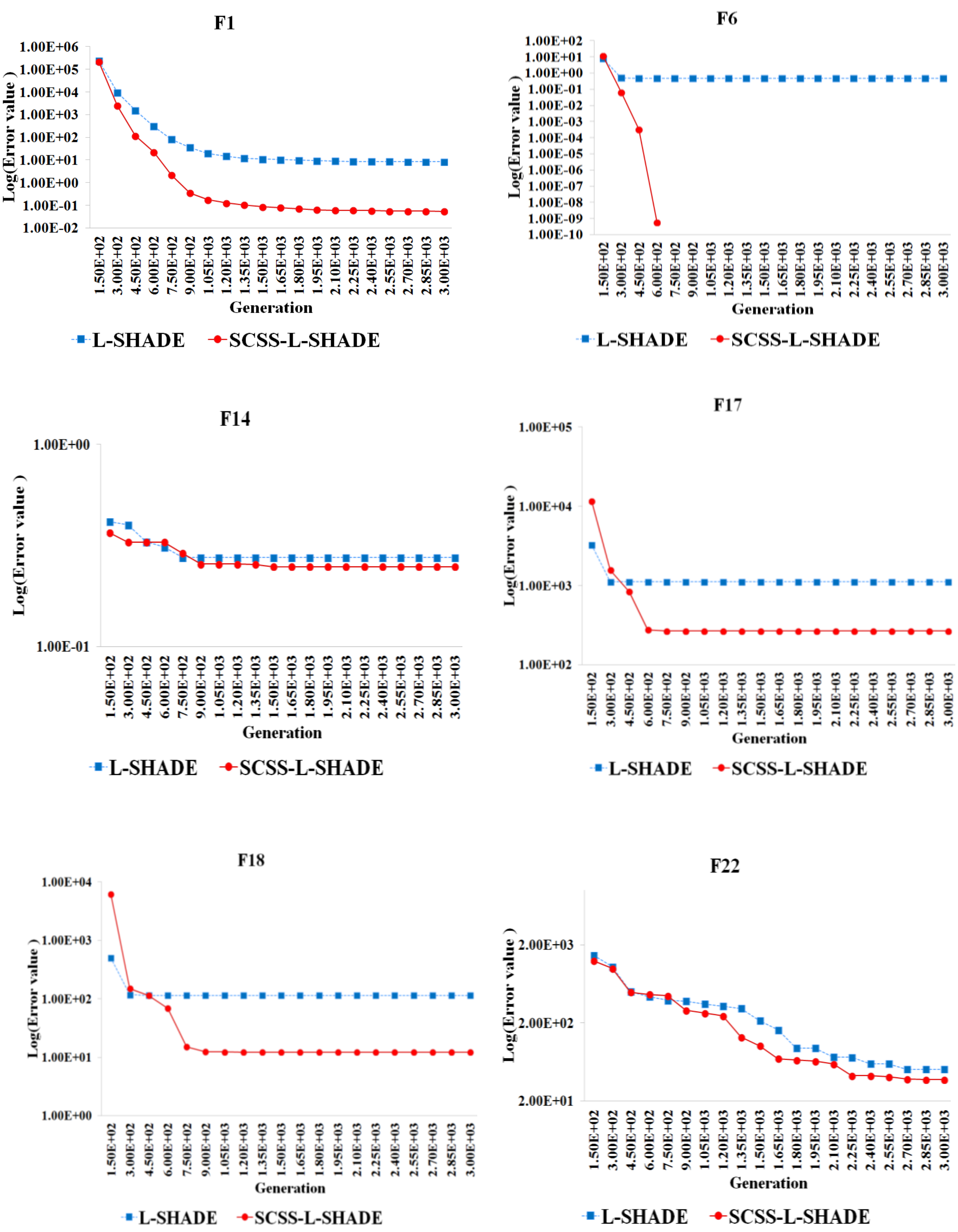

Fig.9 Convergence plot of SCSS-L-SHADE versus L-SHADE on six selected 50-D CEC2014 functions in the median run. (Note: On F6, SCSS-L-SHADE reaches the global optimal at generation 750)

Table 2 Comparison results of SCSS variants with the baselines on CEC2017 test suit

| -/=/+ | 30-D | 50-D |
|---|---|---|
| JADE vs. SCSS-JADE | 19/11/0 | 18/10/2 |
| SHADE vs. SCSS-SHADE | 7/23/0 | 11/19/0 |
| CMA-ES vs. SCSS-CMA-ES | 18/11/1 | 16/14/0 |
| LIPS vs. SCSS-LIPS | 28/1/1 | 28/1/1 |
| L-SHADE vs. SCSS- L-SHADE | 9/18/3 | 15/15/0 |
| UMOEA-II vs. SCSS-UMOEA-II | 3/24/3 | 14/14/2 |
| L-SHADE_EpSin vs. SCSS-L-SHADE_EpSin | 7/21/2 | 13/17/0 |
| jSO vs. SCSS-jSO | 7/23/0 | 12/18/0 |
| Total | 225/240/15 | |

To study scalability, the SCSS framework is also tested on 100-*D* CEC2017 functions. The four top methods are selected for the experiment. As shown in Table S14 and Fig. 10, SCSS still yields remarkable performance improvements on the higher dimensional functions, which are much more difficult than the lower dimensional ones. In the total of 120 cases, SCSS outperforms in 70 (=20+14+16+20) cases and underperforms in 6 (=2+2+0+2) cases. These improvements are attributed to the balanced exploitation and exploration maintained by the SS rule.

Furthermore, the overall performances of the considered algorithms are compared according to multiple problem Wilcoxon's test [50] and Friedman's test [50]. Based on multiple problems Wilcoxon's test, Table 3 shows that the SCSS variants perform significantly better than the corresponding baselines at $\alpha = 0.05$. With respect to the Friedman's test, Table 4 indicates that SCSS-jSO is the best-performing algorithm, which achieves the smallest ranking value of 2.76, followed by SCSS-L-SHADE_EpSin.

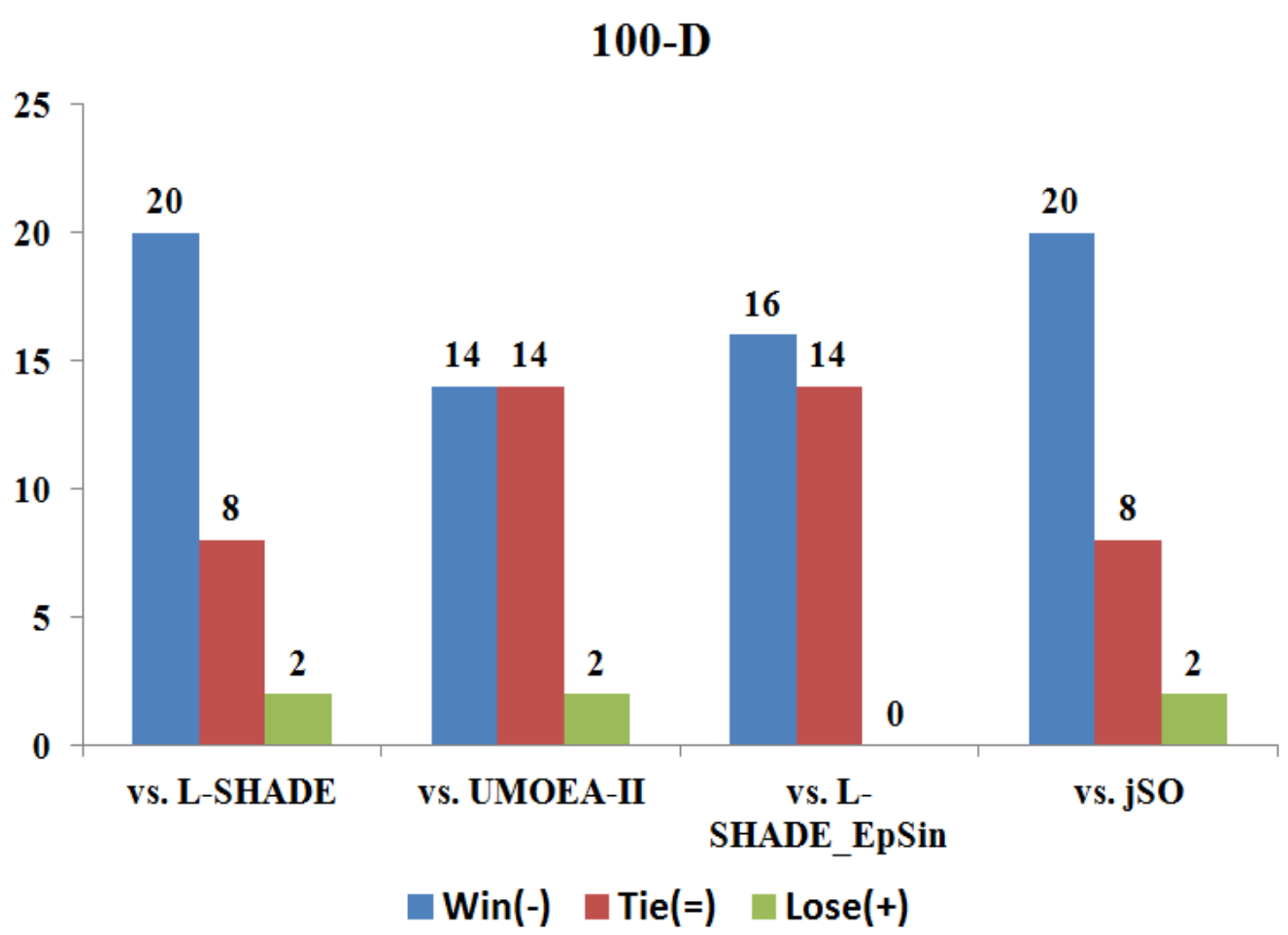

Fig.10 Comparison results of four SCSS-based top algorithms with the baselines on 100-D CEC2017 test functions.

Table 3 Comparison results of the top SCSS variants with the baselines on 30-D, 50-D and 100-D CEC2017 benchmark set according to multi-problem Wilcoxon's test

| | *R*+ | *R*- | *p*-value | $\alpha = 0.05$ |
|---|---|---|---|---|
| SCSS-L-SHADE vs. L-SHADE | 3235.0 | 770.0 | **0.0E+00** | **Yes** |
| SCSS-UMOEA-II vs. UMOEA-II | 3052.5 | 952.5 | **1.7E-05** | **Yes** |
| SCSS-L-SHADE_EpSin vs. L-SHADE_EpSin | 3077.0 | 1018.0 | **3.4E-05** | **Yes** |
| SCSS-jSO vs. jSO | 3710.5 | 384.5 | **0.0E+00** | **Yes** |

Table 4 Overall performance ranking of the considered algorithms on 30-D, 50-D and 100-D CEC2017 benchmark set by Friedman's test

| Algorithm | Ranking |
|---|---|
| SCSS-jSO | **2.76** |
| SCSS-L-SHADE_EpSin | 3.60 |
| jSO | 3.88 |
| L-SHADE_EpSin | 4.33 |
| SCSS-L-SHADE | 4.57 |
| SCSS-UMOEA-II | 5.19 |
| L-SHADE | 5.67 |
| UMOEA-II | 5.96 |

### 3.7 Application in Real-world Problems

We have also applied the proposed method to 22 real-world applications, from [51], where detailed descriptions and source codes of the problems are available. These problems come from various scientific and engineering fields, such as frequency-modulated (FM) sound waves parameter estimation problem, Lennard-Jones potential problem, spread spectrum radar polyphase code design problem, large scale transmission pricing problem and so on [51]. They have a wide range of dimensionality from one up to 216 and are very challenging. As an example, we focused on SCSS-L-SHADE and L-SHADE. Each algorithm has 30 trials with each trial assigned $10^4 \times D$ function evaluations. Note that the maximum function evaluations are kept the same as that used for the previous benchmark functions because higher dimensional functions are more difficult and should be given more resources.

Table 5 tabulates the mean and standard deviations of the solution error values. As shown, SCSS-L-SHADE performs significantly better on 11 problems (including P1, P5-P7, P9, P11, P12, P16-P18 and P20) and loses on none. This demonstrates the reliable performance of SCSS when incorporated with L-SHADE for real-world applications.

Table 5 Performance comparisons (mean (std)) of SCSS-L-SHADE with L-SHADE on 22 CEC2011 real-world problems

| | L-SHADE | SCSS-L-SHADE | | L-SHADE | SCSS-L-SHADE |
|---|---|---|---|---|---|
| ${}_{cec11}P1$ | 0.73 (2.73) − | **0.34 (1.86)** | ${}_{cec11}P12$ | 1050159.77 (1254.39) − | **1047950.07 (1191.33)** |
| ${}_{cec11}P2$ | -27.68 (0.38) = | -27.79 (0.54) | ${}_{cec11}P13$ | 15444.51 (1.56) = | 15444.19 (0.00) |
| ${}_{cec11}P3$ | 0.00 (0.00) = | 0.00 (0.00) | ${}_{cec11}P14$ | 18093.89 (33.47) = | 18093.73 (33.53) |
| ${}_{cec11}P4$ | 18.98 (3.09) = | 17.69 (3.34) | ${}_{cec11}P15$ | 32740.43 (0.21) = | 32740.41 (0.18) |
| ${}_{cec11}P5$ | -36.84 (0.02) − | **-36.82 (0.16)** | ${}_{cec11}P16$ | 123355.03 (580.33) − | **123000.40 (381.84)** |
| ${}_{cec11}P6$ | -29.16555 (0.00) − | **-29.16598 (0.00)** | ${}_{cec11}P17$ | 1735648.35 (7377.90) − | **1729536.58 (5961.35)** |
| ${}_{cec11}P7$ | 1.16 (0.07) − | **1.11 (0.09)** | ${}_{cec11}P18$ | 925951.66 (758.39) − | **925373.83 (489.51)** |
| ${}_{cec11}P8$ | 220.00 (0.00) = | 220.00 (0.00) | ${}_{cec11}P19$ | 934334.22 (700.86) = | 934138.43 (617.75) |
| ${}_{cec11}P9$ | 369.60 (125.46) − | **292.23 (104.70)** | ${}_{cec11}P20$ | 926086.29 (462.05) − | **925719.66 (674.85)** |
| ${}_{cec11}P10$ | -21.60 (0.11) = | -21.62 (0.08) | ${}_{cec11}P21$ | 15.50 (0.57) = | 15.50 (0.62) |
| ${}_{cec11}P11$ | 48154.11(369.11) − | **47274.03 (410.89)** | ${}_{cec11}P22$ | 14.54 (2.40) = | 14.09 (3.05) |
| -/=/+ | **11/11/0** | | | | |

## 4 Conclusion

To address the potential adverse effect of randomness in evolutionary algorithms, a selective-candidate framework with similarity selection rule (SCSS) is proposed in this paper. In SCSS, each current solution owns a pool of $M$ candidates generated by $M$ reproduction procedures. The final candidate is then determined from the pool by a similarity selection method, which is designed based on fitness ranking and Euclidian distance measures. We have described the motivation of the design (Section 2.2.2), incorporated the design into several classic, advanced and top algorithms from EA and SI families (Sections 3.1, 3.2, 3.5 and 3.6), analyzed its working mechanism (Sections 3.3 and 3.4) and have also applied it to solve 22 real-world problems (Section 3.7). Comprehensive experiments show that 1) SCSS significantly enhances the performance of the algorithms under consideration; 2) Scheme 2 performs consistently well, especially on the advanced and top algorithms and is thus recommended; 3) the appropriate $M$ value is relatively small (2 to 4) for the advanced and top algorithms with balanced EEC. According to Section 3.4, $M$ = 2 should be the first choice when testing SCSS in a new metaheuristic since it always brings improvements. One may then further increase $M$ to see whether better performance can be achieved.

The supplementary file and MATLAB codes of SCSS can be downloaded from https://zsxhomepage.github.io/.

## Acknowledgements

The work was supported in part by City University of Hong Kong under a Research Grant (Project No: 7004710) and in part by National Natural Science Foundation of China (No. 61671485).